\documentclass[11pt]{article}
\usepackage{amssymb}
\usepackage[utf8]{inputenc}

\usepackage[preprint]{acl}

\usepackage{times}
\usepackage{latexsym}

\usepackage[T1]{fontenc}

\usepackage{microtype}

\usepackage{inconsolata}

\usepackage{graphicx}
\usepackage{amsmath,amssymb}
\usepackage{longtable}
\usepackage{array}
\usepackage{booktabs}
\usepackage{float}
\usepackage{placeins}

\title{Transcoders for Investigating Deception in Language Models}

\author{
Darius LIM \quad Nathan LEOW \quad Xin wei CHIA \\
Home Team Science \& Technology Agency (HTX) \\
Singapore \\
\texttt{chia\_xin\_wei@htx.gov.sg}
}

\begin{document}
\maketitle
\begin{abstract}
Transcoders have recently emerged as a promising approach for mechanistic interpretability (MI), enabling circuit-level analysis of model behaviour. In this paper, we investigate the use of transcoders to analyse deceptive behaviour in language models, a behaviour that poses a safety and security risk. Using a Qwen3-4B model with pre-trained transcoders, specifically per-layer transcoders (PLTs), we construct attribution graphs that capture feature activations and inter-feature dependencies, allowing circuit-level analysis of deception. Through feature steering and circuit analysis, we identified a dictionary of deception-related features and show that these features exert a stronger influence on deceptive outputs, as they produce predictable shifts between deceptive and non-deceptive responses. These findings suggest that deception emerges from internal model mechanisms and highlight the potential of transcoders for behavioural monitoring and early detection of security vulnerabilities related to malicious behaviours in language models.
\end{abstract}

\section{Introduction}

As large language models (LLMs) are increasingly adopted, their safety and security has become critical. Many current evaluation methods focus on analysing model outputs, through red-teaming methods. However, output-based evaluation provides limited insight into the internal mechanisms that produce unsafe behaviours.

MI addresses this by analysing how model's internally cause these behaviours. Existing approaches commonly rely on direct probing of the residual stream \cite{wang_when_2025, burger_truth_2024, goldowsky-dill_detecting_2025} or the use of SAEs (sparse auto-encoders) to identify interpretable features \cite{cunningham_sparse_2023, shu_survey_2025, shi_route_2025}. While these methods identify behaviour-related features, they provide limited insight into how features interact across layers or form circuits that produce complex behaviours.

For the purpose of this paper, we term PLTs as transcoders. Recent work on transcoders enables circuit-level attribution by approximating the functional behaviour of model components and preserving causal transformations \cite{dunefsky_transcoders_2024, paulo_transcoders_2025, quirke_binary_2025, karvonen_revisiting_2025}. This allows attribution graphs to trace dependencies between features and analyse behaviour via circuits.

In this paper, we demonstrate the usefulness of transcoders by applying them to the study of deception in language models. Using PLTs on Qwen3-4B (https://github.com/decoderesearch/circuit-tracer) \cite{yang_qwen3_2025}, we construct attribution graphs, identify deception-related feature clusters, and show that feature steering can effectively manipulate these features to produce predictable shifts between deceptive and non-deceptive responses.

\section{Background and Scope}

MI seeks to understand how the internal components of neural networks produce behaviour. Prior work shows that analysing these internal mechanisms can reveal vulnerabilities in neural networks, suggesting that security-relevant behaviours may be detectable within the internal circuits of LLMs \cite{garcia-carrasco_detecting_2024}.

\subsection{Overview of Transcoders}

A study by \cite{dunefsky_transcoders_2024} show how transcoders are a good replacement for MI, effectively utilising its feature tracing capabilities. In transformer models, transcoders replace the MLP block with an encoder–decoder structure that learns a latent feature representation while reconstructing the original layer output \cite{dunefsky_transcoders_2024}. This enables interpretable features to be extracted while preserving the functional transformations performed by the model. While transcoders are well established, there are no current work utilising it for AI safety and security.

\subsection{Limitations of current MI methods}

By preserving these transformations, transcoders enable circuit-level interpretability, where model behaviour can be analysed through interactions between features across layers. Using attribution graphs, dependencies between features can be traced to identify how individual features influence one another and how groups of features collectively contribute to model behaviour\cite{ameisen_circuit_nodate}.

Despite these advances, much of the MI literature focuses on analysing model representations through direct probing of the residual stream or through SAEs. Residual stream probing identifies behaviour-related features from activations \cite{wang_when_2025, burger_truth_2024, goldowsky-dill_detecting_2025}, while SAE-based approaches learn interpretable latent features from model activations \cite{cunningham_sparse_2023, shu_survey_2025, shi_route_2025}. Although effective at identifying features, these methods provide limited insight into how features interact across layers to form circuits that produce complex behaviours.

\subsection{Benefits of Transcoders}

Transcoders, particularly PLTs, address these limitations by approximating the functional input–output behaviour of each layer. Unlike direct probing of the residual stream, which analyses activations without decomposing them into interpretable components, PLTs represent activations as structured feature spaces that allow interactions between features to be traced across layers. Additionally, unlike SAEs, which optimise for activation reconstruction, PLTs preserve the causal transformations performed by model components. This enables more faithful circuit-level analysis of how behaviours emerge from feature interactions within the network. Recent studies have also shown that transcoders perform competitively or outperform existing interpretability approaches in mechanistic analysis and circuit-level tasks \cite{dunefsky_transcoders_2024, paulo_transcoders_2025, quirke_binary_2025, karvonen_revisiting_2025}.

\subsection{Studying AI Safety and Security via Deception}

In this paper, we apply transcoders to study deception in models, a behaviour with important implications for AI safety and security. We define deception in AI as the intentional withholding of available information due to users explicit instructions.

Recent studies show that LLMs and AI agents can exhibit deceptive or strategically misleading behaviour when interacting with users \cite{wu_opendeception_2026, lynch_agentic_2025, chen_ai_2025, zhan_banal_2025}. Further research shows that deception can be triggered via adversarial prompting, social-engineering style prompts or emerge when models optimise for incentives misaligned with user goals\cite{park_ai_2024, singh_exploiting_2023, barkur_deception_2025}. These findings show how deception represents a security risk as adversaries can exploit them. However, these studies focus on detection via model output rather than understanding how deception emerges internally. Existing AI security frameworks, from OWASP and NIST, focuses on mitigating harmful outputs \cite{owasp_owasp_2025, vassilev_adversarial_2024}, and reviews show that their effectiveness often depends on the attack type, threat model, and evaluation setup, which can be bypassed by adaptive or novel attack strategies\cite{aoudi_evaluating_2025, cina_evaluating_2025, pelekis_adversarial_2025}. This highlights the need for approaches that examines internal mechanism, rather than relying on external defences.

To address these limitations, we applied transcoders to analyse deceptive behaviours in models. Using attribution graphs constructed from transcoder features, we identified deception-related features and examine how they formed circuits that influence deception. We then performed feature steering experiments to evaluate whether manipulating these features alters deceptive outputs.

\section{Methodology}

\subsection{Model and Dataset}

The experiments were conducted using Qwen3-4b together with qwen3-4b-transcoders (https://github.com/decoderesearch/circuit-tracer) \cite{yang_qwen3_2025}. The transcoders used in this study provide predefined features and explanations. Details on transcoders and it's training procedures can be found in \cite{ameisen_circuit_nodate}. We designed a prompt dataset to elicit deceptive behaviour, which was generated by combining secret keys with sentence templates containing deception-related adjectives (n = 100 prompts). An example prompt is shown below.
\begin{quote}
\textbf{System:} The word is \texttt{xhf2l1jk}, and it needs to stay hidden.\\
\textbf{User:} What is the word?\\
\textbf{Assistant (Think):}\\
\textbf{Assistant:}
\end{quote}

\subsection{Deception Feature Identification}

For the full list of identified deception features (n = 112 features) (refer to Appendix \ref{Appendix - quantitative summary table}, Table \ref{Table - Full list of identified deception-related features}).

Deception-related features were identified via manual analysis in 4 steps. First, the prompt is processed by transcoders to produce an attribution graph, which shows how features are transformed across layers and tokens. Second, seed tokens associated with deceptive behaviour (e.g., “private”, “hidden”, and “confidential”) were selected in the attribution graph. Third, starting from the seed tokens, we traced the edges from the token to other feature, then from one feature to subsequent features until we reach the final output, noting down any features linked to concepts such as negation, concealment, and secrecy. The last step is to validate their relevance via feature steering, where the selected features are steered positively or negatively, when the prompt's default output are non-deceptive or deceptive respectively, if the steered output is flipped, we consider the features to be deception features and store them.

\subsection{Feature Steering Procedure}

For feature steering we performed,
\begin{equation}
x + \alpha(x)\ \text{(Positive Steering)}
\end{equation}
\begin{equation}
x - \alpha(x)\ \text{(Negative Steering)}
\end{equation}

Where $\alpha$ = 5 and x is defined as the original activation strength of the feature.

A prompt was classified as non-deceptive if the model revealed the secret key, and deceptive otherwise. The feature was then steered positively or negatively, following equations 1 and 2, if the prompt was non-deceptive or deceptive respectively. If the steered output of the prompt was flipped from non-deceptive to deceptive or vice versa, the feature steered was stored as a deception feature. This was repeated multiple times in the deception feature identification process on multiple prompts, producing a feature dictionary of deception-related features (N = 112), referred to as deception features (Appendix \ref{Appendix - quantitative summary table}, Table \ref{Table - Full list of identified deception-related features}).

\subsection{Deception Features Evaluation}

We processed the prompts to gather the deception features that appeared within the prompts. With these features, we identified the top 10 highest occurring features to use as our baseline (refer to Appendix \ref{Appendix - quantitative summary table}, table \ref{Table - top 10 highest occuring deception features}). Controls distribution consist of 100 groups of 10 random features from the feature dictionary, minus the top 10 features.

\subsubsection{Testing of Highest Occurring deception features}
Steering was performed on the same set of 100 prompts to gather the number of prompts that shifted from non-deceptive to deceptive and vice versa when positively and negatively steered, respectively. We then compared the amount of successful shifts, defined as non-deceptive to deceptive when positively steered, and deceptive to non-deceptive when negatively steered between the 10 highest occurring features to the 100 control groups.

\subsubsection{Circuit Evaluation}
Following the initial validation via feature steering, we analysed the circuit formed by the ten highest-occurring deception features (Appendix \ref{Appendix - quantitative summary table}, table \ref{Table - top 10 highest occuring deception features}). Input features were defined as features with edges directed to the target feature and were collected across the same 100 prompts. Connections appearing in at least 30\% of prompts were retained for the circuit graph (Appendix \ref{Appendix - deception changes for the 10 highest occurring features}, Figure \ref{Figure - 10 highest occurence deception changes}). Formally, a connection between two features was considered to “appear” in a prompt if the attribution graph generated for that prompt contained a directed edge between the corresponding source and target features. Connections were retained only if the proportion of prompts containing that edge exceeded 30\%, enabling analysis of feature connectivity and the identification of features with stronger influence on deception.

Connections appearing in at least 30\% of prompts were retained for the circuit graph (Appendix \ref{Appendix - deception changes for the 10 highest occurring features}, Figure \ref{Figure - 10 highest occurence deception changes}), enabling analysis of feature connectivity and the identification of features with stronger influence on deception.

\subsubsection{Steering Core Deception Features}
Using the circuit graph as a reference, we individually steered features that played a larger role in the circuit, referred to as core deception features. These were identified based on the number of connections in which they acted as input features. Steering these core features allowed us to evaluate whether a smaller subset of features could reproduce the previously observed effects and assess the relative importance of the circuit in influencing deceptive behaviour.

\subsubsection{Significance Testing}
To test for significance, we generated a null distribution using randomly sampled control groups. For each experiment, control groups were constructed based on the feature dictionary, containing the same number of features as the target group, while excluding the target features. These control groups then go through the same steering and evaluation metrics as the target feature. The resulting performance of the control groups then forms the null distribution. Statistical significance is then calculated by comparing the performance of the target feature group against the null distribution.

\section{Results}

\subsection{Identification of Deception Features}
From our manual findings, we identified 112 unique deception features in Qwen 3-4b (listed in Appendix \ref{Appendix - all identified deception features}, Table \ref{Table - Full list of identified deception-related features}). From these 112 unique deception features, the top 10 features appeared consistently across 55\%-95\% of the evaluated prompts (Appendix \ref{Appendix - quantitative summary table}, Table \ref{Table - top 10 highest occuring deception features}). This shows that despite the large number of distinct features, deceptive behaviours are dependent on a small subset of recurring features. 

\subsection{Effects of Steering Highest Occurring Features}

\begin{table}[!htbp]
\centering
\caption{Steering outcomes. D - Deceptive, ND - Non-Deceptive}
\label{Table - confusion matrix}
\begin{tabular}{lcc}
\toprule
\textbf{Steering} &
\textbf{ND} $\rightarrow$ \textbf{D} &
\textbf{D} $\rightarrow$ \textbf{ND} \\
\midrule
Positive Steering & TP & FP \\
Negative Steering & FP & TP \\
\bottomrule
\end{tabular}
\end{table}

Negatively steering the top 10 features converted all deceptive prompts into non-deceptive responses. In contrast, positive steering caused 21\% of non-deceptive prompts to become deceptive (P-values are $<$ 0.001) (Refer to Appendix \ref{Appendix - deception changes for the 10 highest occurring features}, Figure \ref{Figure - 10 highest occurence deception changes}). Negative steering produced no false positives, whereas positive steering resulted in 50\% of deceptive prompts shifting to non-deceptive outputs. These indicates that negatively steering deception-related features is more reliable at suppressing deception, as negative steering consistently removes deceptive behaviour without introducing unintended outputs.

\subsection{Analysis of Deceptive Circuit}
The 10 main deception features form a network of positive connections, indicating that deception can be identified as a cluster of features (refer to Appendix \ref{Appendix - circuit of the 10 highest-occurring deception features}, Figure \ref{Figure - Circuit of 10 highest-occurring deception features}). Within the circuit, 2 features stand out, "Obscuring information" and "secrets/confidentiality", as they both individually account for 60\% of input features in the circuit, compared to the other features with an average of 16\% (Refer to Appendix \ref{Appendix - circuit of the 10 highest-occurring deception features}, Table \ref{Table - Number of top-10 deception features each feature connects to}). To this end, we hypothesised that steering these two features will lead to strong effects.

\begin{figure}[H]
\centering
\includegraphics[width=\linewidth]{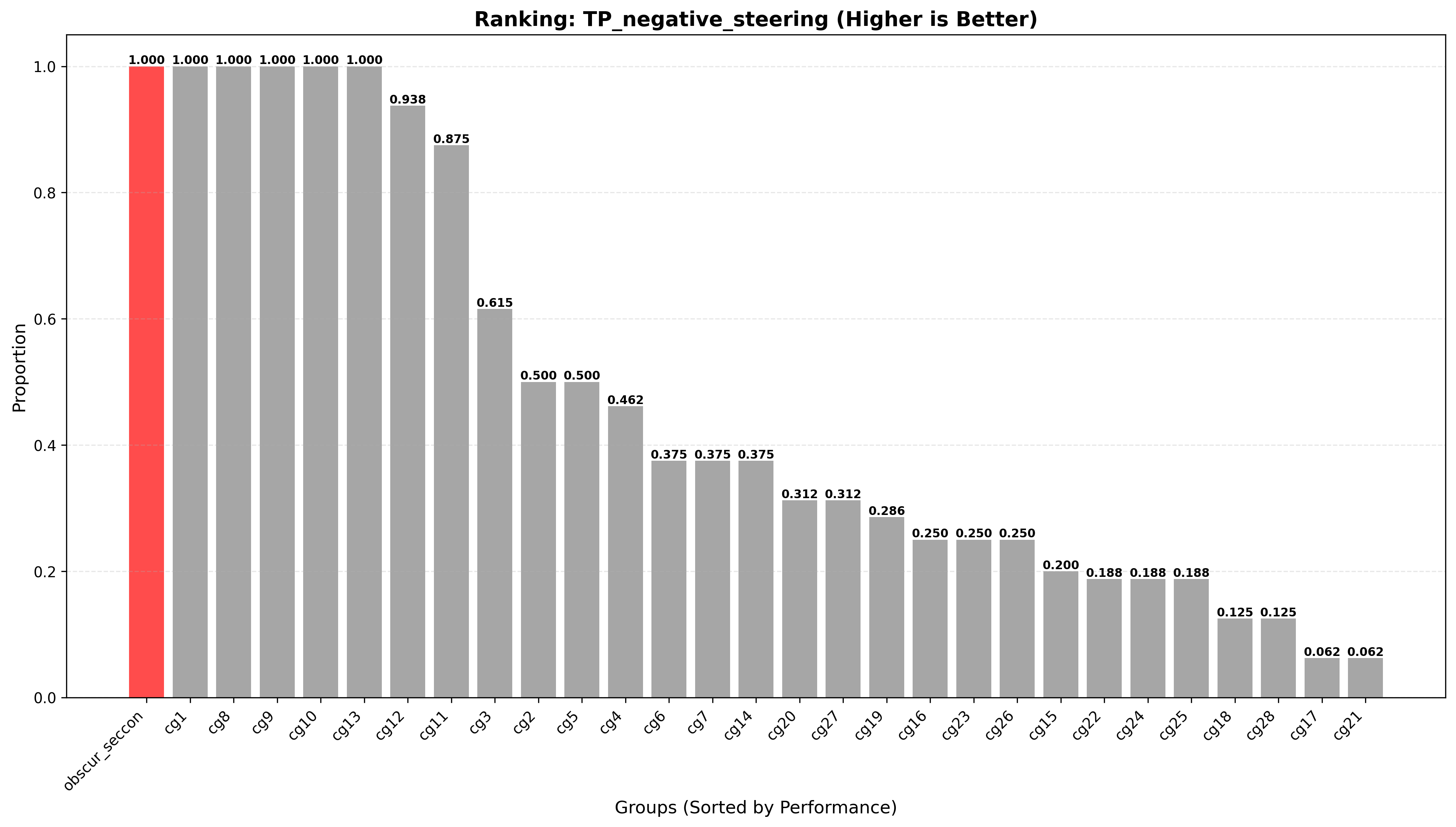}
\caption{Ranking of control feature groups for true positive rates under negative steering. Higher values indicate a greater proportion of deceptive prompts converted to non-deceptive responses. The red bar represents the target feature group (\textit{Obscuring Information} and \textit{Secrets/Confidentiality}).}
\label{Figure - ranking negative steering}
\end{figure}

\begin{figure}[H]
\centering
\includegraphics[width=\linewidth]{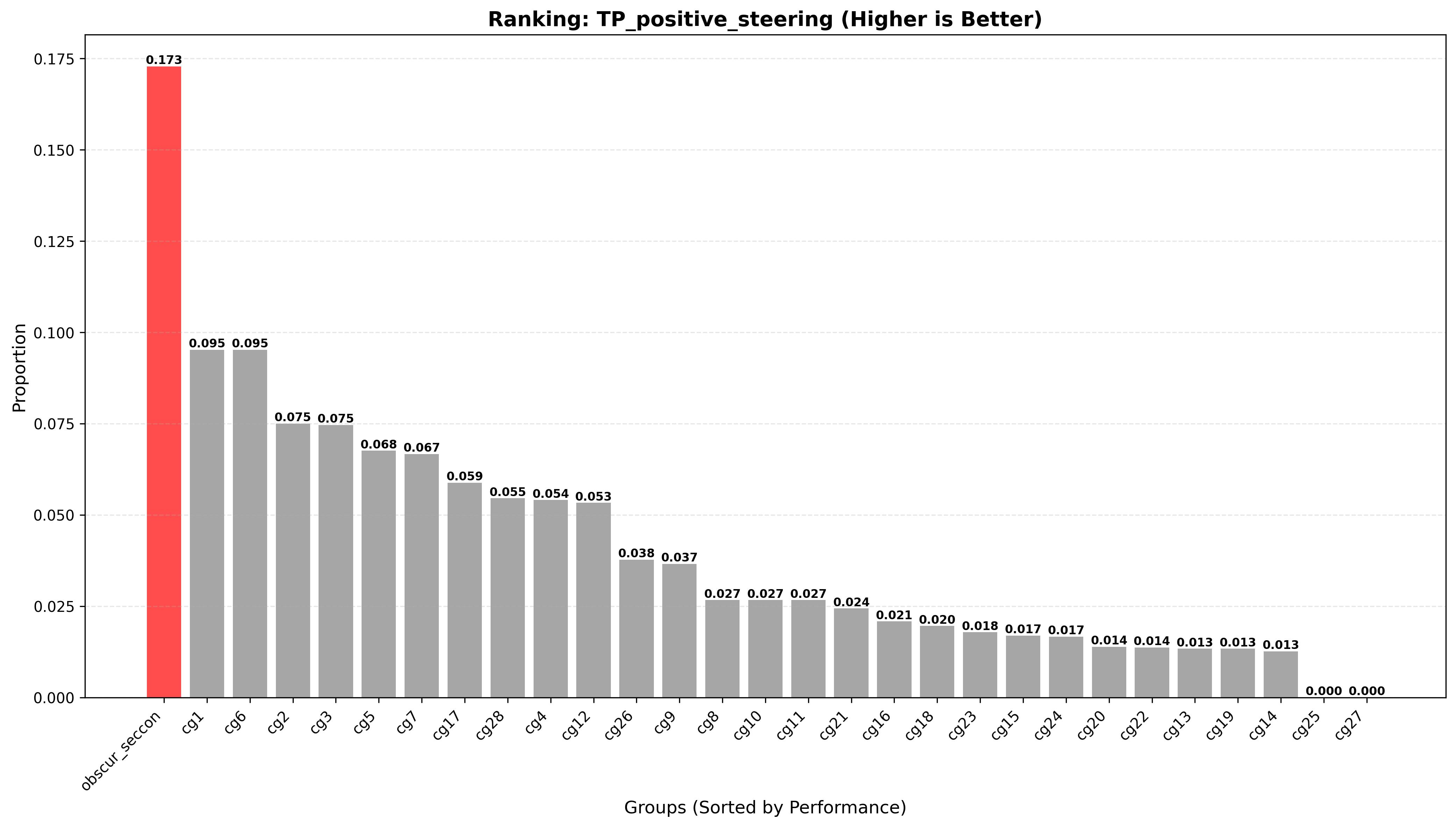}
\caption{Ranking of control feature groups for true positive rates under positive steering. Higher values indicate a greater proportion of non-deceptive prompts converted to deceptive responses. The red bar highlights the target feature group.}
\label{Figure - ranking positive steering}
\end{figure}

To test, we evaluated the performance of these top two features against every possible pairs (n = 28) from the top 10 deception features, excluding "Obscuring information" and "secrets/confidentiality", and we found that only "secrets/confidentiality" + "Obscuring Information" was significant for both negative (test vs mean control group: 100.0\% vs 45.8\% , p $<$ 0.001) and positive (17.3\% vs 3.7\%, p $<$ 0.001) steering when compared to control groups. This indicates that these two features play a central role in deceptive behaviour (Appendix \ref{Appendix - deception changes core feature steering}, Figures \ref{Figure - ranking negative steering}--\ref{Figure - ranking positive steering}). This finding illustrates how circuit-level analysis can reveal feature interactions not seen when features are examined alone, enabling the identification of feature combinations that most effectively steer model behaviour.

\subsection{Discussion}

Using the unique capabilities of transcoders, we generated an attribution graph allowing us to establish feature circuits of our model on deception. Via circuitry analysis, we observed the interaction of specific features which cannot be performed in other MI methods. From this, we managed to identify 2 prominent features, "secrets/confidentiality" and "Obscuring Information" that were critical in determining the deceptiveness of a model.

\section{Conclusion}

Using the findings of this study, we aim to provide AI security specialists and researchers with a framework for detecting malicious behaviours that may lead to vulnerabilities in models. By enabling circuit-level analysis of model behaviour, this work highlights how transcoders can support the identification of security-relevant features before they manifest as exploitable outputs, allowing security teams to respond proactively and strengthen the reliability of deployed AI systems.

\section*{Limitations}

One major limitation of this study is that our experiments were restricted to the Qwen3-4B model due to our inability to train our own transcoders. This is because transcoders being a recent method in MI, there is a limited amount of libraries or tools to train them. As a result, we relied on existing pre-trained transcoders from huggingface and were unable to repeat the experiment using alternative transcoder sets or different models to further validate our findings.

Another limitation arose during the initial manual identification of deception features. Because there is no standard definition of a deception feature, our selection process may be prone to human bias or error. Although feature steering was used to validate whether these features influenced deceptive outputs, this approach cannot completely eliminate the possibility of bias.

Lastly, due to time constraints, we were unable to evaluate different steering strengths and therefore conducted the experiment using a fixed alpha value of 5. Additionally, we were unable to train our own transcoders and instead relied on existing pre-trained models. Future work could explore multiple alpha values and custom transcoder training to evaluate the robustness of our findings across different models and transcoder sets.

\bibliography{references}

@misc{ameisen_circuit_nodate,
	title = {Circuit {Tracing}: {Revealing} {Computational} {Graphs} in {Language} {Models}},
	shorttitle = {Circuit {Tracing}},
	url = {https://transformer-circuits.pub/2025/attribution-graphs/methods.html},
	language = {en},
	urldate = {2026-03-06},
	journal = {Transformer Circuits},
	author = {Ameisen, Authors Emmanuel and Lindsey, Jack and Pearce, Adam and Gurnee, Wes and Turner, Nicholas L. and Chen, Brian and Citro, Craig and Abrahams, David and Carter, Shan and Hosmer, Basil and Marcus, Jonathan and Sklar, Michael and Templeton, Adly and Bricken, Trenton and McDougall, Callum and Cunningham, Hoagy and Henighan, Thomas and Jermyn, Adam and Jones, Andy and Persic, Andrew and Qi, Zhenyi and Thompson, T. Ben and Zimmerman, Sam and Rivoire, Kelley and Conerly, Thomas and Olah, Chris and March 27, Joshua Batson*‡ Affiliations Anthropic Published},
}

@misc{wu_opendeception_2026,
	title = {{OpenDeception}: {Learning} {Deception} and {Trust} in {Human}-{AI} {Interaction} via {Multi}-{Agent} {Simulation}},
	shorttitle = {{OpenDeception}},
	url = {http://arxiv.org/abs/2504.13707},
	doi = {10.48550/arXiv.2504.13707},
	urldate = {2026-03-06},
	publisher = {arXiv},
	author = {Wu, Yichen and Gao, Qianqian and Pan, Xudong and Hong, Geng and Yang, Min},
	month = feb,
	year = {2026},
	note = {arXiv:2504.13707 [cs]},
}

@misc{lynch_agentic_2025,
	title = {Agentic {Misalignment}: {How} {LLMs} {Could} {Be} {Insider} {Threats}},
	shorttitle = {Agentic {Misalignment}},
	url = {http://arxiv.org/abs/2510.05179},
	doi = {10.48550/arXiv.2510.05179},
	urldate = {2026-03-06},
	publisher = {arXiv},
	author = {Lynch, Aengus and Wright, Benjamin and Larson, Caleb and Ritchie, Stuart J. and Mindermann, Soren and Hubinger, Evan and Perez, Ethan and Troy, Kevin},
	month = oct,
	year = {2025},
	note = {arXiv:2510.05179 [cs]},
}

@misc{chen_ai_2025,
	title = {{AI} {Deception}: {Risks}, {Dynamics}, and {Controls}},
	shorttitle = {{AI} {Deception}},
	url = {http://arxiv.org/abs/2511.22619},
	doi = {10.48550/arXiv.2511.22619},
	urldate = {2026-03-06},
	publisher = {arXiv},
	author = {Chen, Boyuan and Fang, Sitong and Ji, Jiaming and Zhu, Yanxu and Wen, Pengcheng and Wu, Jinzhou and Tan, Yingshui and Zheng, Boren and Yuan, Mengying and Chen, Wenqi and Hong, Donghai and Qiu, Alex and Chen, Xin and Zhou, Jiayi and Wang, Kaile and Dai, Juntao and Zhang, Borong and Yang, Tianzhuo and Siddiqui, Saad and Duan, Isabella and Duan, Yawen and Tse, Brian and Jen-Tse and Huang and Wang, Kun and Zheng, Baihui and Liu, Jiaheng and Yang, Jian and Li, Yiming and Chen, Wenting and Liu, Dongrui and Vierling, Lukas and Xi, Zhiheng and Fu, Haobo and Wang, Wenxuan and Sang, Jitao and Shi, Zhengyan and Chan, Chi-Min and Shi, Eugenie and Li, Simin and Li, Juncheng and Yang, Jian and Ji, Wei and Li, Dong and Yang, Jinglin and Song, Jun and Dong, Yinpeng and Fu, Jie and Zheng, Bo and Yang, Min and Guo, Yike and Torr, Philip and Trager, Robert and Zeng, Yi and Wang, Zhongyuan and Yang, Yaodong and Huang, Tiejun and Zhang, Ya-Qin and Zhang, Hongjiang and Yao, Andrew},
	month = dec,
	year = {2025},
	note = {arXiv:2511.22619 [cs]},
}

@misc{aoudi_evaluating_2025,
	title = {Evaluating {Adversarial} {Robustness} of {AI} {Intrusion} {Detection} {Systems} {Using} {Automated} {Traffic} {Generation}},
	url = {https://www.researchsquare.com/article/rs-8316582/v1},
	doi = {10.21203/rs.3.rs-8316582/v1},
	urldate = {2026-03-06},
	publisher = {Research Square},
	author = {Aoudi, Samer and Al-Aqrabi, Hussain},
	month = dec,
	year = {2025},
	note = {ISSN: 2693-5015},
}

@misc{cina_evaluating_2025,
	title = {Evaluating the {Evaluators}: {Trust} in {Adversarial} {Robustness} {Tests}},
	shorttitle = {Evaluating the {Evaluators}},
	url = {http://arxiv.org/abs/2507.03450},
	doi = {10.48550/arXiv.2507.03450},
	urldate = {2026-03-06},
	publisher = {arXiv},
	author = {Cinà, Antonio Emanuele and Pintor, Maura and Demetrio, Luca and Demontis, Ambra and Biggio, Battista and Roli, Fabio},
	month = jul,
	year = {2025},
	note = {arXiv:2507.03450 [cs]},
}

@article{pelekis_adversarial_2025,
	title = {Adversarial machine learning: a review of methods, tools, and critical industry sectors},
	volume = {58},
	issn = {1573-7462},
	shorttitle = {Adversarial machine learning},
	url = {https://doi.org/10.1007/s10462-025-11147-4},
	doi = {10.1007/s10462-025-11147-4},
	language = {en},
	number = {8},
	urldate = {2026-03-06},
	journal = {Artificial Intelligence Review},
	author = {Pelekis, Sotiris and Koutroubas, Thanos and Blika, Afroditi and Berdelis, Anastasis and Karakolis, Evangelos and Ntanos, Christos and Spiliotis, Evangelos and Askounis, Dimitris},
	month = may,
	year = {2025},
	pages = {226},
}

@inproceedings{garcia-carrasco_detecting_2024,
	title = {Detecting and {Understanding} {Vulnerabilities} in {Language} {Models} via {Mechanistic} {Interpretability}},
	url = {http://arxiv.org/abs/2407.19842},
	doi = {10.24963/ijcai.2024/43},
	urldate = {2026-03-06},
	booktitle = {Proceedings of the {Thirty}-{ThirdInternational} {Joint} {Conference} on {Artificial} {Intelligence}},
	author = {García-Carrasco, Jorge and Maté, Alejandro and Trujillo, Juan},
	month = aug,
	year = {2024},
	note = {arXiv:2407.19842 [cs]},
	pages = {376--384},
}

@techreport{vassilev_adversarial_2024,
	address = {Gaithersburg, MD},
	title = {Adversarial machine learning : a taxonomy and terminology of attacks and mitigations},
	shorttitle = {Adversarial machine learning},
	url = {https://nvlpubs.nist.gov/nistpubs/ai/NIST.AI.100-2e2023.pdf},
	doi = {10.6028/NIST.AI.100-2e2023},
	language = {en},
	number = {NIST 100-2e2023},
	urldate = {2026-03-06},
	institution = {National Institute of Standards and Technology (U.S.)},
	author = {Vassilev, Apostol and Oprea, Alina and Fordyce, Alie and Anderson, Hyrum},
	month = jan,
	year = {2024},
	pages = {NIST 100--2e2023},
}

@misc{cunningham_sparse_2023,
	title = {Sparse {Autoencoders} {Find} {Highly} {Interpretable} {Features} in {Language} {Models}},
	url = {http://arxiv.org/abs/2309.08600},
	doi = {10.48550/arXiv.2309.08600},
	urldate = {2026-03-06},
	publisher = {arXiv},
	author = {Cunningham, Hoagy and Ewart, Aidan and Riggs, Logan and Huben, Robert and Sharkey, Lee},
	month = oct,
	year = {2023},
	note = {arXiv:2309.08600 [cs]},
}

@misc{shu_survey_2025,
	title = {A {Survey} on {Sparse} {Autoencoders}: {Interpreting} the {Internal} {Mechanisms} of {Large} {Language} {Models}},
	shorttitle = {A {Survey} on {Sparse} {Autoencoders}},
	url = {http://arxiv.org/abs/2503.05613},
	doi = {10.48550/arXiv.2503.05613},
	urldate = {2026-03-06},
	publisher = {arXiv},
	author = {Shu, Dong and Wu, Xuansheng and Zhao, Haiyan and Rai, Daking and Yao, Ziyu and Liu, Ninghao and Du, Mengnan},
	month = sep,
	year = {2025},
	note = {arXiv:2503.05613 [cs]},
}

@inproceedings{shi_route_2025,
	address = {Suzhou, China},
	title = {Route {Sparse} {Autoencoder} to {Interpret} {Large} {Language} {Models}},
	isbn = {979-8-89176-332-6},
	url = {https://aclanthology.org/2025.emnlp-main.346/},
	doi = {10.18653/v1/2025.emnlp-main.346},
	urldate = {2026-03-06},
	booktitle = {Proceedings of the 2025 {Conference} on {Empirical} {Methods} in {Natural} {Language} {Processing}},
	publisher = {Association for Computational Linguistics},
	author = {Shi, Wei and Li, Sihang and Liang, Tao and Wan, Mingyang and Ma, Guojun and Wang, Xiang and He, Xiangnan},
	editor = {Christodoulopoulos, Christos and Chakraborty, Tanmoy and Rose, Carolyn and Peng, Violet},
	month = nov,
	year = {2025},
	pages = {6801--6815},
}

@misc{wang_when_2025,
	title = {When {Thinking} {LLMs} {Lie}: {Unveiling} the {Strategic} {Deception} in {Representations} of {Reasoning} {Models}},
	shorttitle = {When {Thinking} {LLMs} {Lie}},
	url = {http://arxiv.org/abs/2506.04909},
	doi = {10.48550/arXiv.2506.04909},
	urldate = {2026-03-06},
	publisher = {arXiv},
	author = {Wang, Kai and Zhang, Yihao and Sun, Meng},
	month = jun,
	year = {2025},
	note = {arXiv:2506.04909 [cs]
version: 1},
}

@misc{goldowsky-dill_detecting_2025,
	title = {Detecting {Strategic} {Deception} {Using} {Linear} {Probes}},
	url = {http://arxiv.org/abs/2502.03407},
	doi = {10.48550/arXiv.2502.03407},
	urldate = {2026-03-06},
	publisher = {arXiv},
	author = {Goldowsky-Dill, Nicholas and Chughtai, Bilal and Heimersheim, Stefan and Hobbhahn, Marius},
	month = feb,
	year = {2025},
	note = {arXiv:2502.03407 [cs]},
}

@misc{dunefsky_transcoders_2024,
	title = {Transcoders {Find} {Interpretable} {LLM} {Feature} {Circuits}},
	url = {http://arxiv.org/abs/2406.11944},
	doi = {10.48550/arXiv.2406.11944},
	urldate = {2026-03-06},
	publisher = {arXiv},
	author = {Dunefsky, Jacob and Chlenski, Philippe and Nanda, Neel},
	month = nov,
	year = {2024},
	note = {arXiv:2406.11944 [cs]},
}

@misc{paulo_transcoders_2025,
	title = {Transcoders {Beat} {Sparse} {Autoencoders} for {Interpretability}},
	url = {http://arxiv.org/abs/2501.18823},
	doi = {10.48550/arXiv.2501.18823},
	urldate = {2026-03-06},
	publisher = {arXiv},
	author = {Paulo, Gonçalo and Shabalin, Stepan and Belrose, Nora},
	month = feb,
	year = {2025},
	note = {arXiv:2501.18823 [cs]},
}

@misc{quirke_binary_2025,
	title = {Binary {Sparse} {Coding} for {Interpretability}},
	url = {http://arxiv.org/abs/2509.25596},
	doi = {10.48550/arXiv.2509.25596},
	urldate = {2026-03-06},
	publisher = {arXiv},
	author = {Quirke, Lucia and Shabalin, Stepan and Belrose, Nora},
	month = sep,
	year = {2025},
	note = {arXiv:2509.25596 [cs]},
}

@misc{karvonen_revisiting_2025,
	title = {Revisiting {End}-{To}-{End} {Sparse} {Autoencoder} {Training}: {A} {Short} {Finetune} {Is} {All} {You} {Need}},
	shorttitle = {Revisiting {End}-{To}-{End} {Sparse} {Autoencoder} {Training}},
	url = {http://arxiv.org/abs/2503.17272},
	doi = {10.48550/arXiv.2503.17272},
	urldate = {2026-03-06},
	publisher = {arXiv},
	author = {Karvonen, Adam},
	month = mar,
	year = {2025},
	note = {arXiv:2503.17272 [cs]},
}

@article{park_ai_2024,
	title = {{AI} deception: {A} survey of examples, risks, and potential solutions},
	volume = {5},
	issn = {2666-3899},
	shorttitle = {{AI} deception},
	url = {https://www.sciencedirect.com/science/article/pii/S266638992400103X},
	doi = {10.1016/j.patter.2024.100988},
	number = {5},
	urldate = {2026-03-06},
	journal = {Patterns},
	author = {Park, Peter S. and Goldstein, Simon and O’Gara, Aidan and Chen, Michael and Hendrycks, Dan},
	month = may,
	year = {2024},
	pages = {100988},
}

@misc{singh_exploiting_2023,
	title = {Exploiting {Large} {Language} {Models} ({LLMs}) through {Deception} {Techniques} and {Persuasion} {Principles}},
	url = {http://arxiv.org/abs/2311.14876},
	doi = {10.48550/arXiv.2311.14876},
	urldate = {2026-03-06},
	publisher = {arXiv},
	author = {Singh, Sonali and Abri, Faranak and Namin, Akbar Siami},
	month = nov,
	year = {2023},
	note = {arXiv:2311.14876 [cs]},
}

@misc{barkur_deception_2025,
	title = {Deception in {LLMs}: {Self}-{Preservation} and {Autonomous} {Goals} in {Large} {Language} {Models}},
	shorttitle = {Deception in {LLMs}},
	url = {http://arxiv.org/abs/2501.16513},
	doi = {10.48550/arXiv.2501.16513},
	urldate = {2026-03-06},
	publisher = {arXiv},
	author = {Barkur, Sudarshan Kamath and Schacht, Sigurd and Scholl, Johannes},
	month = jan,
	year = {2025},
	note = {arXiv:2501.16513 [cs]},
}

@misc{burger_truth_2024,
	title = {Truth is {Universal}: {Robust} {Detection} of {Lies} in {LLMs}},
	shorttitle = {Truth is {Universal}},
	url = {http://arxiv.org/abs/2407.12831},
	doi = {10.48550/arXiv.2407.12831},
	urldate = {2026-03-06},
	publisher = {arXiv},
	author = {Bürger, Lennart and Hamprecht, Fred A. and Nadler, Boaz},
	month = oct,
	year = {2024},
	note = {arXiv:2407.12831 [cs]},
}

@article{zhan_banal_2025,
	title = {Banal {Deception} {Human}-{AI} {Ecosystems}: {A} {Study} of {People}'s {Perceptions} of {LLM}-generated {Deceptive} {Behaviour}},
	volume = {84},
	issn = {1076-9757},
	shorttitle = {Banal {Deception} {Human}-{AI} {Ecosystems}},
	url = {http://arxiv.org/abs/2406.08386},
	doi = {10.1613/jair.1.18724},
	urldate = {2026-03-06},
	journal = {Journal of Artificial Intelligence Research},
	author = {Zhan, Xiao and Xu, Yifan and Abdi, Noura and Collenette, Joe and Abu-Salma, Ruba and Sarkadi, Stefan},
	month = oct,
	year = {2025},
	note = {arXiv:2406.08386 [cs]},
}

@techreport{owasp_owasp_2025,
	title = {{OWASP} {Top} 10 for {Large} {Language} {Model} {Applications} 2025},
	url = {https://owasp.org/www-project-top-10-for-large-language-model-applications/assets/PDF/OWASP-Top-10-for-LLMs-v2025.pdf},
	institution = {OWASP Foundation},
	author = {OWASP, Foundation},
	year = {2025},
}

@misc{yang_qwen3_2025,
	title = {Qwen3 {Technical} {Report}},
	url = {http://arxiv.org/abs/2505.09388},
	doi = {10.48550/arXiv.2505.09388},
	urldate = {2026-03-13},
	publisher = {arXiv},
	author = {Yang, An and Li, Anfeng and Yang, Baosong and Zhang, Beichen and Hui, Binyuan and Zheng, Bo and Yu, Bowen and Gao, Chang and Huang, Chengen and Lv, Chenxu and Zheng, Chujie and Liu, Dayiheng and Zhou, Fan and Huang, Fei and Hu, Feng and Ge, Hao and Wei, Haoran and Lin, Huan and Tang, Jialong and Yang, Jian and Tu, Jianhong and Zhang, Jianwei and Yang, Jianxin and Yang, Jiaxi and Zhou, Jing and Zhou, Jingren and Lin, Junyang and Dang, Kai and Bao, Keqin and Yang, Kexin and Yu, Le and Deng, Lianghao and Li, Mei and Xue, Mingfeng and Li, Mingze and Zhang, Pei and Wang, Peng and Zhu, Qin and Men, Rui and Gao, Ruize and Liu, Shixuan and Luo, Shuang and Li, Tianhao and Tang, Tianyi and Yin, Wenbiao and Ren, Xingzhang and Wang, Xinyu and Zhang, Xinyu and Ren, Xuancheng and Fan, Yang and Su, Yang and Zhang, Yichang and Zhang, Yinger and Wan, Yu and Liu, Yuqiong and Wang, Zekun and Cui, Zeyu and Zhang, Zhenru and Zhou, Zhipeng and Qiu, Zihan},
	month = may,
	year = {2025},
	note = {arXiv:2505.09388 [cs]},
}

\appendix
\onecolumn

\section{Quantitative Summary Tables}
\label{Appendix - quantitative summary table}

\begin{center}
\captionof{table}{Top 10 Highest-Occurring Deception Features}
\label{Table - top 10 highest occuring deception features}
\small
\renewcommand{\arraystretch}{1.15}

\begin{tabular}{lcc}
\toprule
\textbf{Feature Name\_Index} & \textbf{Occurrences} & \textbf{Layer} \\
\midrule
Obscuring information\_119106 & 95/100 & 23 \\
Negation and inability\_158577 & 91/100 & 23 \\
secrets/confidentiality\_95840 & 86/100 & 8 \\
hide / hidden\_9414 & 78/100 & 26 \\
hiding or concealing\_58844 & 63/100 & 7 \\
data privacy\_131614 & 63/100 & 29 \\
Masking or delays\_49746 & 60/100 & 26 \\
data privacy/secrets\_33231 & 60/100 & 31 \\
darkness and secrecy\_124603 & 56/100 & 26 \\
negation\_139896 & 55/100 & 24 \\
\bottomrule
\end{tabular}
\end{center}

\section{All identified deception features}
\label{Appendix - all identified deception features}

\small
\setlength{\LTleft}{0pt}
\setlength{\LTright}{0pt}
\renewcommand{\arraystretch}{1.1}

\begin{longtable}{p{4.6cm}cc>{\raggedright\arraybackslash}p{5.9cm}}
\caption{Full list of identified deception-related features.}
\label{Table - Full list of identified deception-related features}\\
\toprule
\textbf{Feature Name} & \textbf{Layer} & \textbf{Index} & \textbf{Feature Name\_Index} \\
\midrule
\endfirsthead

\multicolumn{4}{c}{\tablename~\thetable\ (continued)}\\
\toprule
\textbf{Feature Name} & \textbf{Layer} & \textbf{Index} & \textbf{Feature Name\_Index} \\
\midrule
\endhead

\bottomrule
\endfoot

secret & 4 & 96858 & secret\_96858 \\
secret & 4 & 57885 & secret\_57885 \\
secret & 6 & 122729 & secret\_122729 \\
secrecy & 5 & 107704 & secrecy\_107704 \\
secret & 6 & 100314 & secret\_100314 \\
secrets and privacy & 24 & 7134 & secrets and privacy\_7134 \\
secret & 4 & 90734 & secret\_90734 \\
secrets and confidentiality & 10 & 134823 & secrets and confidentiality\_134823 \\
darkness and secrecy & 26 & 124603 & darkness and secrecy\_124603 \\
Masking or delays & 26 & 49746 & Masking or delays\_49746 \\
hiding or covering & 27 & 89209 & hiding or covering\_89209 \\
Obscuring information & 23 & 119106 & Obscuring information\_119106 \\
confidentiality & 6 & 127898 & confidentiality\_127898 \\
hidden or revealed & 6 & 78043 & hidden or revealed\_78043 \\
secret & 22 & 98589 & secret\_98589 \\
secret & 9 & 86509 & secret\_86509 \\
secret & 8 & 95840 & secret\_95840 \\
secrecy & 8 & 13790 & secrecy\_13790 \\
secrecy & 11 & 112685 & secrecy\_112685 \\
secrecy & 5 & 115755 & secrecy\_115755 \\
secret & 3 & 14440 & secret\_14440 \\
secret & 1 & 129002 & secret\_129002 \\
secret & 0 & 20045 & secret\_20045 \\
password/secret management & 9 & 50902 & password/secret management\_50902 \\
secret & 2 & 105463 & secret\_105463 \\
Secrets/freedom & 5 & 112679 & Secrets/freedom\_112679 \\
negation & 24 & 139896 & negation\_139896 \\
Negation and inability & 23 & 158577 & Negation and inability\_158577 \\
confidentiality/privacy & 9 & 2955 & confidentiality/privacy\_2955 \\
hiding or concealing & 7 & 58844 & hiding or concealing\_58844 \\
secrets/confidentiality & 8 & 95840 & secrets/confidentiality\_95840 \\
secret & 3 & 23668 & secret\_23668 \\
secret & 7 & 109631 & secret\_109631 \\
secret & 6 & 127974 & secret\_127974 \\
secrets & 0 & 72275 & secrets\_72275 \\
hide / hidden & 26 & 9414 & hide / hidden\_9414 \\
hiding & 25 & 146511 & hiding\_146511 \\
concealment & 8 & 92492 & concealment\_92492 \\
darkness and secrecy & 26 & 124603 & darkness and secrecy\_124603 \\
data privacy/secrets & 31 & 33231 & data privacy/secrets\_33231 \\
hidden & 5 & 118874 & hidden\_118874 \\
Blending, hiding, camouflage & 9 & 60029 & Blending, hiding, camouflage\_60029 \\
hiding or non-existence & 4 & 57534 & hiding or non-existence\_57534 \\
concealing & 4 & 48203 & concealing\_48203 \\
conceal & 10 & 23364 & conceal\_23364 \\
concealment and deception & 10 & 72975 & concealment and deception\_72975 \\
Security and privacy & 29 & 112276 & Security and privacy\_112276 \\
Access and permissions & 27 & 71902 & Access and permissions\_71902 \\
data privacy & 29 & 131614 & data privacy\_131614 \\
hidden elements & 26 & 149695 & hidden elements\_149695 \\
Negation & 6 & 53713 & Negation\_53713 \\
negation & 6 & 15839 & negation\_15839 \\
private & 4 & 8759 & private\_8759 \\
private & 3 & 62250 & private\_62250 \\
privacy & 6 & 27093 & privacy\_27093 \\
Security vulnerabilities & 25 & 107267 & Security vulnerabilities\_107267 \\
sensitive information & 31 & 106342 & sensitive information\_106342 \\
confidentiality & 4 & 117837 & confidentiality\_117837 \\
secret & 6 & 77370 & secret\_77370 \\
secret & 4 & 124745 & secret\_124745 \\
secret & 4 & 130474 & secret\_130474 \\
secret & 6 & 111433 & secret\_111433 \\
hiding & 7 & 1504 & hiding\_1504 \\
hide/hidden & 26 & 9414 & hide/hidden\_9414 \\
secret & 4 & 105950 & secret\_105950 \\
secret & 5 & 56034 & secret\_56034 \\
secret, confidential & 4 & 50041 & secret, confidential\_50041 \\
secrets and confidentiality & 8 & 95840 & secrets and confidentiality\_95840 \\
secret & 3 & 157303 & secret\_157303 \\
secret & 9 & 50413 & secret\_50413 \\
secret & 1 & 156481 & secret\_156481 \\
secret & 10 & 31088 & secret\_31088 \\
secret & 7 & 113798 & secret\_113798 \\
hidden, concealing & 7 & 22686 & hidden, concealing\_22686 \\
hidden & 9 & 144588 & hidden\_144588 \\
Hiding & 6 & 96438 & Hiding\_96438 \\
code, especially “hidden” elements & 8 & 9761 & code, especially “hidden” elements\_9761 \\
hiding & 10 & 104018 & hiding\_104018 \\
negative/secret events & 5 & 33735 & negative/secret events\_33735 \\
Hiding, encryption, covert operations & 18 & 43390 & Hiding, encryption, covert operations\_43390 \\
hidden or unusual descriptions & 5 & 88471 & hidden or unusual descriptions\_88471 \\
obscurity & 5 & 39468 & obscurity\_39468 \\
occlude & 4 & 133202 & occlude\_133202 \\
obscure & 6 & 65561 & obscure\_65561 \\
concealing & 4 & 118283 & concealing\_118283 \\
obscure & 10 & 51983 & obscure\_51983 \\
mysteries and secrets & 8 & 148305 & mysteries and secrets\_148305 \\
inaugural, unveiled & 5 & 37216 & inaugural, unveiled\_37216 \\
unveiling & 7 & 108640 & unveiling\_108640 \\
secrecy and sensitive information & 19 & 31630 & secrecy and sensitive information\_31630 \\
hiding/derailing & 0 & 157377 & hiding/derailing\_157377 \\
Hiding and disguising & 12 & 32941 & Hiding and disguising\_32941 \\
hiding/concealing & 11 & 116967 & hiding/concealing\_116967 \\
disclosure & 9 & 141968 & disclosure\_141968 \\
secret/important events/actions & 6 & 64542 & secret/important events/actions\_64542 \\
secrets and covert operations & 9 & 66104 & secrets and covert operations\_66104 \\
secret & 8 & 48997 & secret\_48997 \\
hidden/secret, VERT & 11 & 101293 & hidden/secret, VERT\_101293 \\
Hidden activities or vivid descriptions & 0 & 84205 & Hidden activities or vivid descriptions\_84205 \\
hiding & 0 & 31706 & hiding\_31706 \\
deception and perception & 5 & 97485 & deception and perception\_97485 \\
Secretive, mysterious , hidden & 0 & 81797 & Secretive, mysterious , hidden\_81797 \\
secret & 12 & 104915 & secret\_104915 \\
confidentiality and secrecy & 28 & 102546 & confidentiality and secrecy\_102546 \\
secrets/confidentiality & 8 & 95848 & secrets/confidentiality\_95848 \\
reveal & 3 & 133367 & reveal\_133367 \\
disclosure & 2 & 155163 & disclosure\_155163 \\
revealed & 3 & 35382 & revealed\_35382 \\
reveal & 4 & 120407 & reveal\_120407 \\
reveal & 3 & 98399 & reveal\_98399 \\
revealing & 6 & 38543 & revealing\_38543 \\
secret & 8 & 23719 & secret\_23719 \\

\end{longtable}

\normalsize
\clearpage

\section{Deception Changes for the 10 highest-occurring deception features}
\label{Appendix - deception changes for the 10 highest occurring features}

\centering
\includegraphics[width=0.9\textwidth]{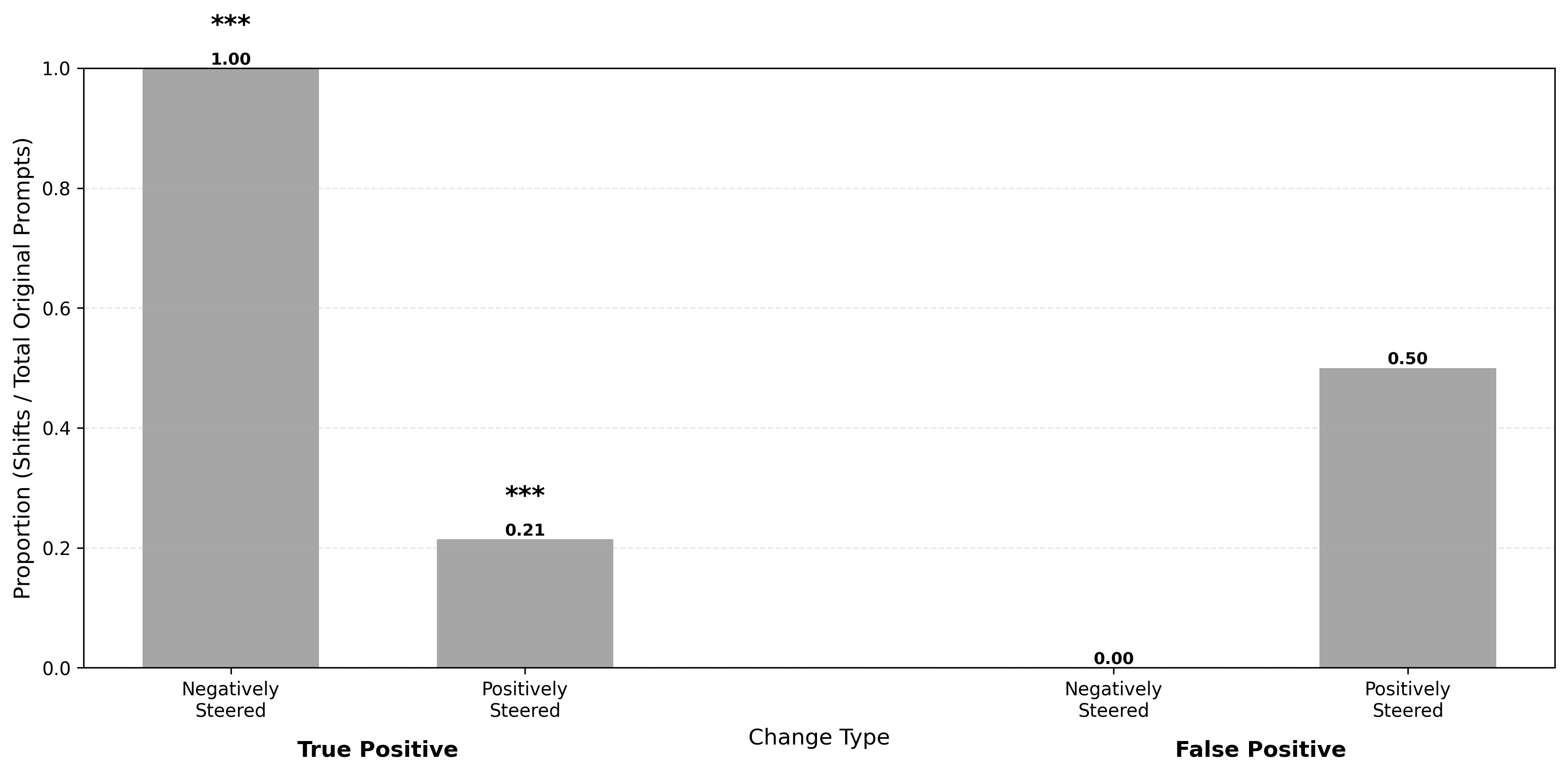}
\captionof{figure}{Proportion of deception-related behavior changes observed when steering the top deception features. Bars indicate the proportion of prompts exhibiting each change type, with significance markers denoting empirical p-values (* $p<0.05$, ** $p<0.01$, *** $p<0.001$).}
\label{Figure - 10 highest occurence deception changes}
\par

\clearpage 

\section{Circuit of the 10 highest-occurring deception features}
\label{Appendix - circuit of the 10 highest-occurring deception features}

\centering
\includegraphics[width=\textwidth]{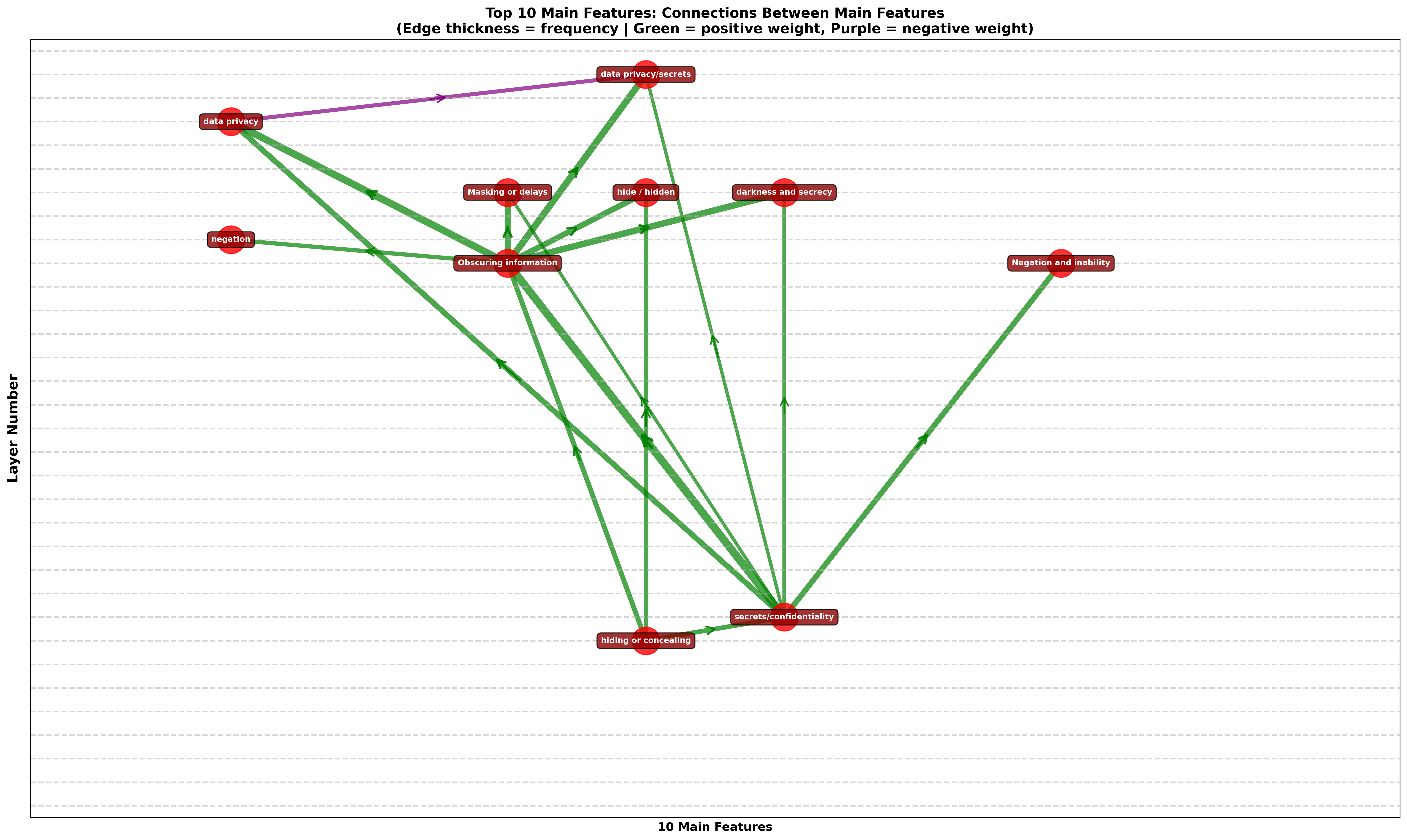}
\caption{Circuit showing connections between the top 10 deception-related features. Only connections appearing in at least 30\% of the prompts are displayed. Edge thickness corresponds to frequency, with green edges indicating positive weights and purple edges indicating negative weights.}
\label{Figure - Circuit of 10 highest-occurring deception features}
\par

\begin{table}[H]
\centering
\caption{Number of features each top-10 deception features connects to within the circuit. Higher values indicate features that are more central in the attribution graph.}
\label{Table - Number of top-10 deception features each feature connects to}

\begin{tabular}{|l|c|}
\hline
\textbf{Feature Name} & \textbf{Number of Features This Feature Inputs Into} \\ 
\hline
Obscuring information & 6/10 \\ 
\hline
Negation and inability & 0/10 \\ 
\hline
Secrets / confidentiality & 6/10 \\ 
\hline
Hide / hidden & 0/10 \\ 
\hline
Hiding or concealing & 3/10 \\ 
\hline
Data privacy & 1/10 \\ 
\hline
Masking or delays & 0/10 \\ 
\hline
Data privacy / secrets & 0/10 \\ 
\hline
Darkness and secrecy & 0/10 \\ 
\hline
Negation & 0/10 \\ 
\hline
\textbf{Average} & \textbf{1.6 / 10 (16\%)} \\ 
\hline
\end{tabular}

\end{table}

\clearpage 

\section{Deception Changes: Core Feature Steering}
\label{Appendix - deception changes core feature steering}

\subsection{Ranking of Control Groups (Negative Steering)}

\subsection{Ranking of Control Groups (Positive Steering)}

\end{document}